\documentclass[runningheads]{llncs}
\usepackage{graphicx}

\usepackage{tikz}
\usepackage{comment}
\usepackage{amsmath,amssymb} % define this before the line numbering.
\usepackage{color}

\usepackage{graphicx}
\usepackage{amsmath,amssymb} % define this before the line numbering.
\usepackage{color}

\long\def\ignore#1{} % RDZ's favorite macro

\usepackage{booktabs}
\usepackage{etoolbox}
\usepackage{epsfig}
\usepackage{graphicx}
\usepackage{amsmath}
\usepackage{amssymb}
\usepackage{mathtools}
\usepackage{pgfplotstable}
\usepackage{pgfplots}
\usepackage[utf8]{inputenc}
\usepackage[english]{babel}
\usepackage[sort, nocompress]{cite}
\usepackage{intcalc}
\usepackage{ifthen}
\usepackage{subcaption}
\usepackage{pgfplots}
\usepackage{booktabs}
\usepackage{floatrow}
\usepackage{pbox}

\definecolor{darkgreen}{RGB}{0, 128, 0}
\definecolor{darkred}{RGB}{128, 0, 0}

\usetikzlibrary{positioning}

\pgfplotsset{
    discard if not/.style 2 args={
        x filter/.code={
            \edef\tempa{\thisrow{#1}}
            \edef\tempb{#2}
            \ifx\tempa\tempb
            \else
                
            \fi
        }
    }
}

\tikzset{
  pics/layers/.style args={#1,#2,#3,#4}{
    code = {
      \def\layers{10}
      \def\layerspacingx{#2}
      \def\layerspacingy{0.1}
      \definecolor{layercolor}{HTML}{DDDDFF}
      \definecolor{darkcolor}{HTML}{888888}
      \foreach \r in {\layers,...,0}
      \ifthenelse{\intcalcMod{\r}{#3}=1}
      {
       \draw [fill=darkcolor, draw=black] (\r*\layerspacingx, \r*\layerspacingy) rectangle ++(#1,#1)
      }
      {
       \draw [fill=#4, draw=black] (\r*\layerspacingx,\r*\layerspacingy) rectangle ++(#1,#1)
      };
    }
  }
}

\pgfplotsset{compat=1.9}

\def\LGate{\mathcal{L}_{\text{G}}}

\long\def\ignore#1{} % RDZ's favorite macro

\usepackage[pagebackref=true,breaklinks=true,letterpaper=true,colorlinks,bookmarks=false]{hyperref}

\begin{document}
% \linenumbers
\pagestyle{headings}
\mainmatter
\def\ECCVSubNumber{5695}  % Insert your submission number here

\title{Channel selection using Gumbel Softmax} % Replace with your title

\begin{comment}
\titlerunning{ECCV-20 submission ID \ECCVSubNumber} 
\authorrunning{ECCV-20 submission ID \ECCVSubNumber} 
\author{Anonymous ECCV submission}
\institute{Paper ID \ECCVSubNumber}
\end{comment}
%******************

%\begin{comment}
\titlerunning{Channel selection using Gumbel Softmax}
\author{Charles Herrmann\inst{1}\orcidID{0000-0002-9576-9394} \and
Richard Strong Bowen\index{Bowen,Richard Strong}\inst{1}\orcidID{0000-0002-9628-5471} \and
Ramin Zabih\inst{1,2}\orcidID{0000-0001-8769-5666}}
% To Springer editor: in "Richard Strong Bowen", Richard is the first name, Strong is the middle name and Bowen is the last name.

%
\authorrunning{C. Herrmann et al.}
\institute{
%Cornell Tech, New York NY 10044, USA\\
Cornell Tech \\
\email{\{cih, rsb, rdz\}@cs.cornell.edu}\\
\and
Google Research\\
%\email{raminz@google.com}
}
%\end{comment}
%******************
\maketitle

\begin{abstract}
\ignore{Important applications such as mobile computing require reducing the  computational costs of neural network inference. Ideally, applications would specify their preferred tradeoff between accuracy and speed, and the network would optimize this end-to-end, using classification error to remove parts of the network \cite{lecun1990optimal,mozer1989skeletonization,BMVC2016_104}. Increasing speed can be done either during training -- e.g., pruning filters \cite{li2016pruning} -- or during inference  -- e.g., conditionally executing a subset of the layers \cite{aig}. We propose a single end-to-end framework that can improve  inference efficiency in both settings. We use a combination of batch activation loss and classification loss, and Gumbel reparameterization to learn network structure \cite{aig,jang2016categorical}. We train end-to-end, and the same technique supports pruning as well as conditional computation. We obtain promising experimental results for ImageNet classification with ResNet \cite{he2016resnet} (45-52\% less computation).}
Important applications such as mobile computing require reducing the  computational costs of neural network inference. Ideally, applications would specify their preferred tradeoff between accuracy and speed, and the network would optimize this end-to-end, using classification error to remove parts of the network. Increasing speed can be done either during training -- e.g., pruning filters -- or during inference  -- e.g., conditionally executing a subset of the layers. We propose a single end-to-end framework that can improve  inference efficiency in both settings. We use a combination of batch activation loss and classification loss, and Gumbel reparameterization to learn network structure. We train end-to-end, and the same technique supports pruning as well as conditional computation. We obtain promising experimental results for ImageNet classification with ResNet  (45-52\% less computation).
\end{abstract}

\section{Pruning and conditional computation}

Despite their great success \cite{alexnet, he2016resnet,Simonyan14c}, convolutional networks remain too computationally expensive for many important tasks. Modern architectures often struggle to run on standard desktop hardware, let alone mobile devices. These computational requirements pose a serious obstacle in settings constrained by latency, power, memory and/or compute; key examples include smartphones, robotics and autonomous driving.  Considerable work has been put into exploring the tradeoffs between computation and performance. Popular approaches include expert-designed efficient networks like MobileNetV2\cite{sandler2018mobilenetv2}, and reinforcement learning to search for more efficient architectures \cite{zoph2018learning,he2018amc}.

We focus on two longstanding lines of research: pruning \cite{lecun1990optimal,mozer1989skeletonization} and conditional computation \cite{Bengio:TR13, aig}. Pruning, in its earliest~\cite{lecun1990optimal,mozer1989skeletonization} and modern~\cite{huang2017data,han2015deep} forms, attempts to remove the least useful parts of the network. The goal is to leave a smaller network with comparable or better accuracy. A network with conditional computation runs lightweight tests that can choose to bypass larger blocks of computation that are not useful for the given input. Aside from benefits in inference-time efficiency~\cite{Bengio:TR13}, skipping computations can improve training time or test performance \cite{figurnov2017spatially,stochasticdepth,aig}, and can provide insight into network behavior \cite{stochasticdepth,aig}. 

Our goal is to improve a neural network by trading off classification error and computation. End-to-end training is a key advantage of neural nets \cite{LeCun:nature15}, but poses a technical challenge. Both pruning and conditional computation are {categorical} decisions which are not easy to optimize by gradient descent. However, Gumbel-Softmax (GS) \cite{BengioLC13,gumbel1954statistical,jang2016categorical,maddison2016concrete} gives a way to address this challenge.

\begin{figure}[h]
    \centering
    \begin{tikzpicture}
    \pic (L1) at (0,2) {layers={1,0.03,1,layercolor}};
    \pic (mask) at (0.25,0.25) {layers={0.2,0.03,4,white}};
    \pic (L2) at (3,1) {layers={1,0.03,4,layercolor}};
    \node[draw] (combine) at (2, 1.75) {$\times$};
    \path [->] (1.4,1) edge[bend right] (combine.south);
    \path [->] (1.4,2.5) edge[bend left] (combine.north);
    \draw [->] (combine.east) -- ++(0.5,0);
    \end{tikzpicture}
    \caption{%Pruning and conditional computation of channels. 
    The clean set of filter outputs (top left) are multiplied channel-wise by a vector of binary random variables (bottom left), which is  
    % Zeros in the binary vector result in some channels being pruned. 
    learned during training. 
    % which effectively performs architecture search in the space of channel subsets. 
    For conditional computation, the gating vector's entries depend upon the input at this layer, while for pruning they do not.
    % so a filter is computed on only certain inputs.
    \label{fig:arch}}
\end{figure}
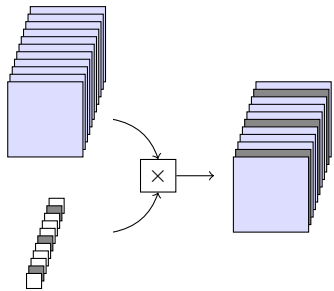

We focus on the ResNet \cite{he2016resnet} architecture, as it is the mainstay of current deep learning techniques for image classification. The general architecture of a prunable channel in a network is shown in Figure~\ref{fig:arch}. The computation of a channel can potentially be skipped by sampling the gating vector of random variables. The associated probabilities are learned during training. Their distributions can be either depend on the layer's input, in which case we perform conditional computation, or be independent, in which case we perform pruning.

We propose a per-batch activation loss function, which allows the network to flexibly avoid computing certain filters and their resulting channels. This in turn supports useful tradeoffs between accuracy and inference speed. Per-batch activation loss, in combination with the Gumbel straight-through trick\cite{jang2016categorical}, encourages the gating vector's probabilities to \emph{polarize}, that is, move towards 0 or 1. Polarization has proved to be beneficial 
% for both conditional computation and pruning 
\cite{BMVC2016_104,courbariaux2015binaryconnect}.
% TODO: move below into a later section?

We summarize our contributions as follows:
\begin{itemize}
    \item We explore conditional computation at the channel level and significantly outperform other techniques.\ignore{, improving the level of specialization displayed by the network. }
    \item We investigate the use of Gumbel soft-max for pruning a network at the channel-level in an end-to-end manner. We  identify a mathematical property of the combination of our batch activation loss and Gumbel soft-max that encourages polarization.
    \item We demonstrate that a single technique can achieve significant results in both areas.
\end{itemize}

This paper is organized as follows. We begin by introducing notation and reviewing related work. Section~\ref{sec:contrib} introduces and analyses our per-batch activation loss function and inference strategies, and discusses the role of polarization. Experimental results on ImageNet and CIFAR-10 are presented in Section~\ref{sec:data}, for both conditional computation and for pruning. Our best experimental results for pruning reduce computation by 51\% on ResNet; our best results for conditional computation, reduce computation on ImageNet by 45--52\% on ResNet.
Additional experiments and more details are included in the supplemental material.

\subsection{Gating neural networks}
\label{sec:math}

In order to learn a discrete structure such as a network architecture with the continuous method of stochastic gradient descent, we learn a probability distribution over structures, and minimize the expected loss. In this work, we learn whether or not to compute a channel. Let $\mathcal G$ be a set of gates indexed by $i$:

\begin{itemize}
    \item $Z_i$, a 0-1 random variable which is $1$ with probability $p_i$.
    \item $g_i$, a portion of the network which computes $p_i$.
\end{itemize}

 When $g$ and thus $Z$ also depend on the input image $j$ we write $g_{ij}$, $p_{ij}$, and $Z_{ij}$. Where $g_i$ depends on the input we use the phrase ``data-dependent''; where $g_i$ does not, ``data-independent''. We use Gumbel Softmax and straight-through training \cite{gal2017concrete,jang2016categorical} to train $g_i$. To generate the vector of $Z_i$s, we run each $g_i$ and then sample. If $Z_i=0$, the associated filter is not run, we simply replace the corresponding channel with a block of zeros. We use the straight-through trick: at training time during the forward pass, we use $Z_i$ and during back-propagation, we treat $Z_i$ as $p_i$. We define the ``activation rate'' of the batch as $\frac{1}{|\mathcal G||\mathcal B|}\sum_{0\leq i\leq |\mathcal G|}\sum_{0\leq j\leq |\mathcal B|} Z_{i,j}$ where $\mathcal B$ is the batch of inputs the network sees. This captures the fraction of the channels being computed for all gates over a batch. The ``activation rate'' of a gate $i$ is  $\frac{1}{|\mathcal B|}\sum_{0\leq j\leq |\mathcal B|} Z_{i,j}$
This captures the fraction of time that the channel $i$ is computed for the given batch.

\subsection{Our loss}

The intuition behind our loss is that we want to encourage the activation rate for each batch to approach a target rate hyperparameter $t$. Smaller values of $t$ will correspond to less computation.
Our batch activation loss is defined as
\begin{equation}\label{eq:batchloss}
\mathcal{L}_B = \left ( t - \frac{1}{|\mathcal B| |\mathcal G|}\sum_{0\leq i \leq |G|} \sum_{0 \leq j \leq  |\mathcal B|}  Z_{i,j}\right )^2 
\end{equation}

\section{Related work}

Our technique allows us to learn a network with conditional computation (using data-dependent gates), or a smaller, pruned network (using data-independent gates). As such, we describe our relation to both fields, as well as related work on regularization.

\subsection{Conditional computation}

%Conditional computation has been well studied in computer vision. 
Cascaded classifiers \cite{viola2004robust} shorten computation by identifying easy negatives and have been adapted to deep learning \cite{li2015convolutional,yang2016exploit}. 
More recently, \cite{msdnet} and \cite{mcgill2017deciding} both propose a cascading architecture which computes features at multiple scales and allows for dynamic evaluation, where at inference time the user can tradeoff speed for accuracy. Similarly, \cite{branchnet} adds intermediate classifiers and returns a label once the network reaches a specified confidence. \cite{figurnov2017spatially,Graves2016AdaptiveCT} both use the state of the network to adaptively decrease the number of computational steps during inference. \cite{Graves2016AdaptiveCT} uses an intermediate state sequence and a halting unit to limit the number of blocks that can be executed in an RNN; \cite{figurnov2017spatially} learns an image dependent stopping condition for each ResNet block that conditionally bypasses the rest of the layers in the block. \cite{shazeer2017outrageously} trains a large number of small networks and then uses gates to select a sparse combination for a given input. \cite{bolukbasi2017adaptive} selects the most-efficient network for a given input and also uses early-exit.

The most closely related work is AIG \cite{aig}, which probabilistically gates individual layers during both training and inference, with a data-dependent gating computation. The major difference between AIG and our work is that they \emph{specify} target rates for each gate, whereas we \emph{learn} these values, by giving a target rate for the entire network. The reason for this difference is that AIG focuses on inducing specialization on the network, whereas we focus on improving the run-time of these networks. This focus on specialization is reflected in their loss, which has a target rate $t_i$ for each gate $i$: $\LGate{} = \frac{1}{|\mathcal{G}|} \sum_{0 \leq i < |\mathcal{G}|}\left ( t_i - \frac{1}{|\mathcal{B}|} \sum_{0 \leq j < |\mathcal{B}|} Z_{i,j} \right )^2$

As reported in their code\footnote{See
\url{https://github.com/andreasveit/convnet-aig}}, the target rates for each layer of ResNet-50 are $[1$, $1$, $0.8$, $1$, $t$, $t$, $t$, $1$, $t$, $t$, $t$, $t$, $t$, $1$, $0.7$, $1]$ for $t\in [0.4, 0.5, 0.6, 0.7]$. This loss function forces specialization since each gate learns to run at its target rate. However, constraining each gate identically is inflexible; for example, consider a dataset with two labels that are equally distributed. If the target rate $t$ is different than $0.5$, no layer will easily specialize to one of the labels. This value of $t$ also determines the approximate speed of the final conditional network; networks trained with $t=0.5$ will be about twice as fast as the baseline network. Yet AIG will push every layer with target rate $t$ to specialize to run on half the data. This rules out many possible network configurations. Ideally, we want to pass in a single global target rate $t$ for the network's speed and then allow the network to learn the optimal distribution of data for its gates. It can then choose to specialize individual gates on the the subsets which benefit the most from additional computation, and not be constrained to the gate's target rate.

In addition, the loss that AIG uses cannot be adapted to network pruning, since it does not allow any the activation rate of any gate to approach 0 or 1 (a gate turning completely on or off). Additionally, modern network pruning is done on a channel-basis, which increases the number of gates from 17 for their layer version of ResNet-50 to thousands of gates for the channels of ResNet-18.

In summary, our loss enables the following improvements vis-a-vis AIG:

\begin{itemize}
    \item Support for pruning. AIG only supports conditional computation\ignore{ since each gate has its own target rate}.
    \item More granular control. AIG specifies per-gate target rates. Specifying per-gate target rates is infeasible at the scale of channels. Instead, our approach learns a rate for each gate, given a soft constraint on the full network.
    \item Improved performance. Our loss function gives the network more flexibility to configure the activation rates of individual gates. We find experimentally that our network can take advantage of this flexibility to make very different gate assignments (as demonstrated by comparing our Figure \ref{fig:heatmap} and AIG's Figure~4). We also produce a much lower FLOPs count with comparable accuracy (as shown in Table \ref{tab:conditional}).
\end{itemize}

We provide an experimental comparison with AIG in Section~\ref{sec:data} and an ablation comparison in Section~\ref{secsub:ablationfiltervslayer}.

\subsection{Pruning}

Network pruning is another approach to decreasing computation time. Researchers initially attempted to determine the importance of specific weights \cite{hassibi1993second,lecun1990optimal} or hidden units \cite{mozer1989skeletonization} and remove those which are unimportant or redundant. Weight-based pruning on CNNs follows the same fundamental approach; \cite{han2015learning} prunes weights with small magnitude and \cite{han2015deep} incorporates these into a pipeline which also includes quantization and Huffman coding. Numerous techniques prune at the channel level, whether through heuristics \cite{hu2016network,li2016pruning, he2019filter} or approximations to importance \cite{he2017channel, molchanov2016pruning, suau2018principal}. \cite{luo2017thinet} prunes using statistics from the following layer. \cite{topruneornottoprune} applies gates to a layer's weight tensors, sorts the weights during train time, and then sends the lowest to zero. Contemporary with our work, \cite{you2019gate} uses a Taylor expansion rather than Gumbel to estimate the impact of opening or closing a gate; their technique prunes the network, but has no natural extension to conditional computation.  Additionally, \cite{liu2018rethinking} suggests that the main benefits of pruning come primarily from the identified architecture. 

Recently, several attempts have been made at doing channel-based pruning in an end-to-end manner. \cite{huang2017data} adds sparsity regularization and then modifies stochastic Accelerated Proximal Gradient to prune the network end-to-end. Our work differs from \cite{huang2017data} by using Gumbel Softmax to integrate the sparsity constraint into an additive loss which can be trained by any optimization technique; we use unmodified stochastic gradient descent with momentum (SGD), the standard technique for training classification.

Similarly, \cite{luo2018autopruner} uses the per-batch results of each layer to learn a per-layer ``code''. These codes are then used to learn a mask for the layer. As training progresses, these masks are driven to be $0-1$ by increasing a sigmoid temperature term. The term in their loss function which trades off against computation time is similar to our per-batch activation loss defined in Equation~\ref{eq:batchloss}. Their architecture does not use stochasticity or the Gumbel trick; we do not use a similar sigmoid temperature term, because we find that the variance term implicit in the loss is sufficient for pruning. See Section~\ref{sec:contrib} for more details. We also provide an experimental comparison in Section~\ref{sec:data}.

\ignore{is the main benefit from pruning was performing an architecture search. Following this, we describe our approach as a search over a specific architecture space.
}

\subsection{Regularization and architecture search}

Several regularization techniques, such as Dropout\cite{dropout} and Stochastic Depth\cite{stochasticdepth}, have explored gating different parts of the network to make the final network more robust and less prone to over-fitting. Both techniques try to induce redundancy through probabilistically removing parts of the network during training. Dropout ignores individual units and Stochastic Depth skips entire layers. 
%Both provide evidence that the increased redundancy helps to prevent over-fitting. 
These techniques can be seen as gating units or layers, respectively, where the gate probabilities are hyperparameters.

%louizos2017learning
In the Bayesian machine learning community, data-independent gating is used as both a form of regularization and for architecture search. Their regularization approaches can be seen as generalizing dropout by learning the dropout rates. \cite{BMVC2016_104}~performs pruning by learning multipliers for weights, which are encouraged to be $0-1$ by a sparsity-inducing loss $w(1-w)$. \cite{gal2017concrete}~proposes per-weight regularization, using the straight-through Gumbel-Softmax trick. \cite{srinivas2016generalized}~uses a form of trainable dropout, learning a per-neuron gating probability. 
%These are regularized by their likelihood against a beta distribution, and training is done with the straight-through trick. 
\cite{Srinivas_2017_CVPR_Workshops}~learns sparsity at the weight level using a binary mask. They adopt a complexity loss ($L_0$ on weights) plus a sparsification loss similar to~\cite{BMVC2016_104}.  
\cite{louizos2017learning}~extends the straight-through trick with a hard sigmoid to obtain less biased estimates of the gradient. They use a loss equal to the sum of Bernoulli weights, which is similar to a per-batch activation loss. \cite{pmlr-v70-molchanov17a}~extends the variational dropout in \cite{NIPS2015_5666}~to allow dropout probabilities greater than a half. 
%Training with the straight-through trick and placing a log-scale uniform prior on the dropout probabilities, they find substantial sparsification with minimal change in change in accuracy, including on some vision problems.

Recently, several techniques have used binary gating or masking terms for architecture search. \cite{shirakawa2018dynamic} uses Bernoulli random variables to dynamically learn network architecture elements, like connectivity, activation functions, and layers. Similarly, \cite{cai2018proxylessnas} learns a gating structure for convolutional blocks of different sizes, pools, etc. and proposes an end-to-end and reinforcement learning approach.

\section{Technical considerations}
\label{sec:contrib}

A number of issues arise when applying our batch activation loss to speed inference. We begin with a discussion of polarization. We then describe training considerations followed by inference strategies. Finally we discuss our overall loss function and how to integrate gates into the ResNet architecture.

\subsection{Gate polarization}
\label{secsub:polarization}

Polarization plays a key role in several respects, and occurs extensively in our experimental results (see \ref{sec:data}). 
In the framework laid out in Section~\ref{sec:math}, the $p_i$ are a mechanism for learning discrete structures; in the independent case, a network architecture, and in the dependent case, an adaptive (or per-input) network architecture. The situation where the probabilities polarize corresponds to the continuous mechanism arriving at a discrete answer.

For data-independent gates, polarization corresponds to $Z_i$ collapsing to always be either $0$ or $1$: in other words, each gate permanently chooses to run or skip its respective channel. In a perfectly polarized data-independent gating configuration some channels are never computed, and the network acts as a deterministic, pruned network. For data-dependent gates, polarization does not necessarily imply that the activation rate of a specific gate, $\frac{1}{|\mathcal B|}\sum_{0\leq j\leq |\mathcal B|} Z_{i,j}$ is $0$ or $1$; just that  $\forall j$, $Z_{i,j}$ is either $0$ or $1$; under polarization, a gate's activation rate can have any value between $0$ and $1$. Conceptually, polarization means that for a given input, the decision whether or not to compute the channel is deterministic.

We observe that our batch activation loss has a property that actively encourages polarization in the independent case. Since $\mathcal L_B$ is a random variable, SGD and the straight-through trick can be seen as minimizing its expected value~\cite{jang2016categorical}.

\begin{property}
In the independent case, the expected batch activation loss is $0$ only if each $g_i$ is polarized.
\end{property}
To see why this property holds, note that the expected activation loss is

%[{\mathcalL}_B] = \left (t - \mathbb{E}\left [ \frac{1}{|\mathcal B| |\mathcal G|} \sum_{0 \leq j < |\mathcal B|} \sum_{ 0 \leq i < |\mathcal G|} Z_{i,j} \right ] \right )^2
%\mathbb{E}[{\mathcalL}_B] = \left (t - \mathbb{E}\left [ \frac{1}{|\mathcal B| |\mathcal G|} \sum_{0 \leq j < |\mathcal B|} \sum_{ 0 \leq i < |\mathcal G|} Z_{i,j} \right ] \right )^2 \\
% \mathbb{E}[{\mathcal L}_B] &= \left (t - \mathbb{E}\left [ \frac{1}{|\mathcal B| |\mathcal G|} \sum_{0 \leq j < |\mathcal B|} \sum_{ 0 \leq i < |\mathcal G|} Z_{i,j} \right ] \right )^2 \\
%&+\textrm{Var} \left (\frac{1}{|\mathcal B| |\mathcal G|} \sum_{0 \leq j < |\mathcal B|} \sum_{0 \leq i < |\mathcal G|} Z_{i,j} \right)

% \left (t - \mathbb{E}\left [ \frac{1}{|\mathcal B| |\mathcal G|} \sum_{0 \leq j < |\mathcal B|} \sum_{ 0 \leq i < |\mathcal G|} Z_{i,j} \right ] \right )^2
% \left ( \frac{1}{|{\mathcal{B}||{\mathcal{G}|}\sum_{0 \leq j < |\mathcal{B}|} \sum_{0 \leq i < |\mathcal{G}|} Z_{i,j} \right)
%\vspace{2mm}
%\qquad\qquad\quad$Q = \frac{1}{|\mathcal{B}||\mathcal{G}|}\displaystyle\sum_{0 \leq j < |\mathcal{B}|} \displaystyle\sum_{0 \leq i < |\mathcal{G}|} Z_{i,j}$

\begin{equation*}
    \mathbb{E}[\mathcal{L}_B] = \left(t - \mathbb E\left[Q\right]\right)^2 + \textrm{Var}\left(Q\right) \quad\mathrm{where}\quad Q = \frac{1}{|\mathcal{B}||\mathcal{G}|}\displaystyle\sum_{0 \leq j < |\mathcal{B}|} \displaystyle\sum_{0 \leq i < |\mathcal{G}|} Z_{i,j}
\end{equation*}

Clearly both terms in the expectation are non-negative and the second term (the variance) is only $0$ at polarized values. The first term encourages the overall activation rate of the network to be close to $t$, but allows the  activation rate of individual gates to vary. The second term generally encourages gate polarization.

\subsection{Training considerations}

As written in Equation \ref{eq:batchloss}, $\mathcal L_{B}$ is a random variable which we cannot back-propagate through. To solve this problem, we use the Gumbel reparameterization and straight-through training \cite{gal2017concrete,jang2016categorical} to train the network. We fixed the Gumbel softmax temperature at 1.0. We found that the straight-through trick ($Z_i \in \{0,1\}$) typically had better performance than the soft version (e.g., $Z_i$ being the Gumbel softmax of $(p, 1-p)$). For static pruning, $g_j$ is simply two parameters that do not receive any input from the network and are directly passed to the Gumbel softmax. For dynamic pruning, $g_j$ consists of an average pool across the image dimensions, a 1d convolution, a batch-norm, a ReLu, and then a final 1d convolution.

In image classification, the standard training regime includes global weight decay, which is equivalent to a squared $L_2$ norm on all weights in the network. We now describe an interaction between this regularization and gate polarization, which motivates a scaling of the weight decay parameter.

Generally, the Gumbel softmax trick reparameterizes the choice of a $k$-way categorical variable to a learning $k$ (unnormalized) logits. In the specific $k=2$ case for on-off gates, we learn two logits $w_0$ and $w_1$ for each gate. In the independent case, these two logits are themselves network parameters and therefore subject to weight decay. Given $w_0$ and $w_1$, the gate's on probability is just the sigmoid of their difference $p = \frac{1}{1 + e^{w_0-w_1}}$. The $L_2$ regularization implicitly adds the following to the loss: $\label{eq:weightdecay}
w_0^2 + w_1^2 = \frac{1}{2}\left( (w_0 + w_1)^2 + \ln\left (\frac{1-p}{p}\right )^2\right)
$

The left hand term drives the logits towards $w_0 = -w_1$. We note that the logits $(w,-w)$ can produce any probability $p$. Since we are interested in the effect of weight decay on the learned gate probabilities, we focus primarily on the second term. It has the opposite of a polarizing effect: it reaches its minimum at $p=0.5$. Since the weight decay loss is summed over all gates, this loss increases directly in proportion to the number of gates. We find that a weight decay parameter of $10^{-4}$ is suitable for a network of $10$ to $20$ gates. However, the implicit weight decay loss is a sum over probabilities whereas the variance term (Eq. \ref{eq:batchloss}) is an average. Therefore, we adopted a heuristic rule: for gating parameters, we divide the weight decay coefficient by the number of gates. Although the above analysis applies to the independent case, we found the same rule was effective for the dependent case.

\subsection{Inference strategies}

Once training has produced a deep network with stochastic gates, it is necessary to decide how to perform inference. The simplest approach is to leave the gates in the network and allow them to be stochastic during inference time. This is the technique that AIG uses. Experimentally, we observe a small variance so this may be sufficient for most use cases. One way to take advantage of the stochasticity is to create an ensemble composed of multiple runs with the same network. Then any kind of ensemble technique can be used to combine the different runs: voting, weighing, boosting, etc. In practice, we observe a bump in accuracy from this ensemble technique, though there is obviously a computational penalty.

However, stochasticity has the awkward consequence that multiple classification runs on the same image can return different results. There are several techniques to remove the stochasticity from the network. The gates can be removed, setting $Z=1$ at test time. This is natural when viewing these gates as a regularization technique, and is the technique used by Stochastic Depth and Dropout. Alternately, inference can be made deterministic by using a threshold $\tau$ instead of sampling. Thresholding with value $\tau$ means that a layer will be executed if the learned probability $p_i$ is greater than $\tau$. This also allows the user some  degree of dynamic control over the computational cost of inference. If the user passes in a very high $\tau$, then fewer layers will activate and inference will be faster. In our experiments, we set $\tau=\frac{1}{2}$. Note that we observe polarization for a large number of our per-batch experiments (particularly with data-independent gates). For a wide range of $\tau$, thresholding leaves a network that behaves almost identically to a stochastic network; additionally, for a large number of $\tau$ the behavior of the thresholded network will be the same.

\subsection{Architectural considerations}

\begin{figure}
\centering
\centering
\begin{tikzpicture}[scale=0.5, every node/.style={transform shape}]
\definecolor{opcolor}{HTML}{DDFFFF}
\definecolor{actcolor}{HTML}{FFDDFF}
\definecolor{mulcolor}{HTML}{FFFFDD}
\definecolor{gatecolor}{HTML}{DDDDDD}

% flow
\node[align=center, draw=black, fill=actcolor] (input) {input};
\node[align=center, draw=black, fill=mulcolor, above =0cm of input] (inputmask) {Input mask};
\node[above = of inputmask, align=center, draw=black, fill=opcolor] (conv1) {Conv 1x1, BN, Relu};
\node[align=center, draw=black, fill=mulcolor, above = 0cm of conv1] (expandmask1) {ConvMask1};
\node[above = of expandmask1, align=center, draw=black, fill=opcolor] (conv2) {Conv 3x3 /2, BN, Relu};
\node[align=center, draw=black, fill=mulcolor, above = 0cm of conv2] (expandmask2) {ConvMask2};
\node[above = of expandmask2, align=center, draw=black, fill=opcolor] (conv3) {Conv 1x1, BN, Relu};
\node[align=center, draw=black, fill=mulcolor, above = 0cm of conv3] (outputmask) {Output mask};

% shortcut

\node[right = of conv2, align=center, draw=black, fill=opcolor] (shortcut) {Conv 1x1 /2, BN, Relu};
\node[align=center, draw=black, fill=mulcolor, above = 0cm of shortcut] (shortcutmask) {ShortcutMask};

% gating units

  \node[align=center, draw=black, fill=gatecolor, left = 2cm of inputmask.center] (inputgate) {$g_{\text{inp}}$}; 
  \node[align=center, draw=black, fill=gatecolor, left = 2cm of expandmask1.center] (expandgate1) {$g_{\text{exp1}}$}; \node[align=center, draw=black, fill=gatecolor, left = 2cm of expandmask2.center] (expandgate2) {$g_{\text{exp2}}$}; 
  \node[align=center, draw=black, fill=gatecolor, left = 2cm of outputmask.center] (outputgate) {$g_{\text{out}}$}; 
  \node[align=center, draw=black, fill=gatecolor, right = of outputmask] (shortcutgate) {$g_{\text{shortcut}}$}; 
  
  \node[above = of outputmask, align=center, draw=black, fill=opcolor] (add) {addition};
%arrows

\draw [->] (inputmask) -- (conv1);
\draw [->] (expandmask1) -- (conv2);
\draw [->] (expandmask2) -- (conv3);
\draw [->] (outputmask) -- (add);
\draw [->] (input.east) -| (shortcut.south);
\draw [->] (shortcutmask.north) |- (add.east);

\draw [dashed, ->] (inputgate) -- (inputmask);
\draw [dashed, ->] (expandgate1) -- (expandmask1);
\draw [dashed, ->] (expandgate2) -- (expandmask2);
\draw [dashed, ->] (outputgate) -- (outputmask);
\draw [dashed, ->] (shortcutgate) |- (shortcutmask);

\end{tikzpicture}
\caption{Gating on ResNet Bottleneck Block}
\label{fig:architectures}
\end{figure}
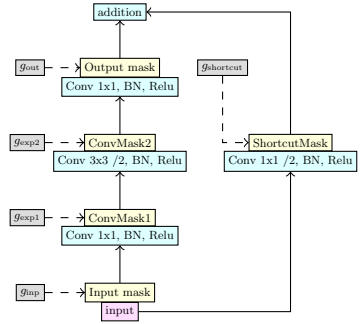

In Figure~\ref{fig:architectures}, we show the blocks for ResNet in its strided form. In Equation \ref{eq:batchloss}, each gate is given equal weight in the activation loss calculation. However for more complex gating schemes, not all gates control the same number of FLOPs (floating point operations per second). To compensate for this, we make a small change to batch activation loss; we change the activation loss to calculate the number of FLOPs using the $Z_{i,j}$. In this case, so $\mathcal L_{\mathcal B}$ takes the following form: $\mathcal L_{\textrm{FLOPs}} = \left(t - \frac{\textrm{\# FLOPs}}{\textrm{Max \# FLOPs}}\right)^2$. Our algorithm is to minimize the sum of this and and classification loss: $\mathcal L = \mathcal L_{\mathcal C} + \mathcal L_{\textrm{FLOPs}}$ where $L_{\mathcal C}$ is the classification loss.

\section{Experiments}
\label{sec:data}

We implemented our method in PyTorch~\cite{pytorch}. Our primary experiments centered around  ResNet\cite{he2016resnet}, running our resulting network on ImageNet\cite{deng2009imagenet}. Our main finding is that our techniques improve both accuracy and inference speed. We also perform an ablation study
% empirical investigation into our networks 
in order to better understand their performance.

\subsection{Training parameters}

For ResNet, we kept the same training schedule as AIG\cite{aig}, and follow the standard ResNet training procedure: batch size of 256, momentum of $0.9$, and weight decay of $10^{-4}$. For the weight decay for gate parameters, we use $\frac{20}{|\mathcal G|}\cdot 10^{-4}$. We train for 100 epochs from a pretrained model of the appropriate architecture with step-wise learning rate starting at $0.1$, and decay by $0.1$ after every $30$ epochs. This is the same training schedule as \cite{aig}. We use standard training data-augmentation (random resize crop to 224, random horizontal flip) and standard validation (resize the images to $256\times256$ followed by a $224\times224$ center crop). \ignore{For input resizing, we follow the TensorFlow ``fast\_mode'' setting\footnote{\url{https://github.com/tensorflow/models/blob/master/research/slim/preprocessing/inception_preprocessing.py}} which uses bilinear resizing at both training and test time.}

In practice, we noticed that many of our ResNet-50 models were not yet at convergence after this training schedule. In order to perform a fair comparison with \cite{aig}, we did not train our data-dependent networks further. For our data-independent networks, we use the same training schedule as ``fine-tune'' in \cite{he2019filter}.

We observe that configurations with low activations rates for gates cause the batch norm estimates of mean and variance to be slightly unstable.\ignore{; for example, if a gate is open 20\% of the time, then the batch norm will be given a abnormal distribution.} Therefore before final evaluation for models trained with smaller batch size, we run training with a learning rate of zero and a large batch size for 200 batches in order to improve the stability and performance of the BatchNorm layers. Unless otherwise specified, we use deterministic inference with a threshold of $0.5$.

\subsection{Results on ImageNet}

A graphical representation of the experimental results are in Figure~\ref{fig:data_graph}, as well as a detailed tabulated breakdown in Tables \ref{tab:pruning} and \ref{tab:conditional}.

\subsubsection{Pruning results}

Results are shown in Figure \ref{fig:resind} and Table \ref{tab:pruning}. We find that we can prune about 43\% of the FLOPs with almost no loss of accuracy from the baseline model. In addition, we can achieve a higher Top-1 accuracy, $76.2$,  with 37\% fewer FLOPs than ResNet-50. Compared to the natural competitor, AutoPruner\footnote{Note that the number we use for their FLOPs is different from what they report. They report lower FLOPs for the baseline ResNet-50 architecture (3.8 GFLOPs versus our 4.028). To normalize the comparison, we added 0.2 GFLOPs to their results.}\cite{luo2018autopruner}, with slightly fewer FLOPs, we have $0.8$ higher accuracy. In additional, we perform nearly 0.7\% better than the best baseline, Filter Pruning via Geometric Median\cite{he2019filter}, with a slightly smaller model (43\% reduction compared to 42\% reduction).

\subsubsection{Conditional computation results}

\begin{table}
\begin{tabular}{|c|ccc|}
\hline
 Model &  \pbox{1.3cm}{\centering Top-1 acc.(\%)} &  \pbox{1.3cm}{\centering Top-5 acc. (\%)} & FLOPs \\ \hline\hline
ANN\cite{bolukbasi2017adaptive} at tradeoff 1 & 74.9 &  91.8  & 2.7\\
ANN\cite{bolukbasi2017adaptive} at tradeoff 2 & 74 & 91.8 & 2.6 \\
%ANN at tradeoff 3 & 71 &  91.7 & 1.9 \\
MSDNet\cite{msdnet}-3.1 & 75.8 & - & 3.1 \\
MSDNet\cite{msdnet}-3 & 75 & - &  3 \\
MSDNet\cite{msdnet}-2.1 & 74 & - &  2.1 \\
%MSDNet-1 & 71 & - & 1 \\
AIG\cite{aig} $t=0.4$ & 75.25 & 92.39 & 2.76\\
AIG\cite{aig} $t=0.5$ & 75.58 & 92.58 & 2.91 \\
AIG\cite{aig} $t=0.6$ & 75.78 & 92.79 & 3.08 \\ 
AIG\cite{aig} $t=0.7$ & 76.18 &  92.92 & 3.26 \\
Ours dep $t=.5$ & \bf76.30 & \bf93.01 & 2.21 \\
Ours dep $t=.4$ & 75.19 & 92.50 & 1.67 \\
Ours dep $t=.3$ & 76.04 & 92.79 & \bf 1.94 \\

 \hline
\end{tabular}
\caption{Comparison of conditional computation on ImageNet-2012.}\label{tab:conditional}
\end{table}

\begin{figure}
%\hspace{-30pt}
\begin{subfigure}[b]{0.45\textwidth}
\begin{tikzpicture}[scale=0.8]
  \begin{axis} [xlabel=FLOPs($\times 10^9$),
                ylabel=ImageNet Top-1 Acc(\%),
                grid=major,
                legend pos = south east]
      \addplot [color=red, mark=*, discard if not={model}{res50dep}] table [x=gflops, y=prec1] {combined.dat};
      \addlegendentry{Ours dep};

      \addplot [color=darkgreen, mark=*, discard if not={model}{aigres50dep}] table [x=gflops, y=prec1] {combined.dat};
      \addlegendentry{AIG\cite{aig}};

      \addplot [color=blue, mark=*, discard if not={model}{msdnet}] table [x=gflops, y=prec1] {combined.dat};
      \addlegendentry{MSDNet\cite{msdnet}};

      \addplot [color=black, mark=*, discard if not={model}{ann}] table [x=gflops, y=prec1] {combined.dat};
      \addlegendentry{ANN\cite{bolukbasi2017adaptive}};

      \addplot [color=cyan, mark=*] coordinates {(4.028, 76.13)};
      \addlegendentry{ResNet-50\cite{he2016resnet}};

      \draw [dashed, color=cyan] (axis cs:\pgfkeysvalueof{/pgfplots/xmin}, 76.13) -- (axis cs:\pgfkeysvalueof{/pgfplots/xmax}, 76.13);
\end{axis}
\end{tikzpicture}
\caption{Conditional computation results for ResNet-50}\label{fig:data_graph}
\label{fig:resdep}
\end{subfigure}
\quad \quad \quad 
%\hspace{25pt}
\begin{subfigure}[b]{0.45\textwidth}
\centering
\begin{tikzpicture}[scale=0.8]
  \begin{axis} [xlabel=FLOPs($\times 10^9$),
                ylabel=ImageNet Top-1 Acc(\%),                xmax=5,
                grid=major,
                legend pos = south east]
      \addplot [color=red, mark=*, discard if not={model}{res50ind}] table [x=gflops, y=prec1] {combined.dat};
      \addlegendentry{Ours ind};

      \addplot [color=darkgreen, mark=*, discard if not={model}{dds}] table [x=gflops, y=prec1] {combined.dat};
      \addlegendentry{DDS\cite{huang2017data}};

      \addplot [color=blue, mark=*, discard if not={model}{autopruner}] table [x=gflops, y=prec1] {combined.dat};
      \addlegendentry{AutoPruner\cite{luo2018autopruner}};

      \addplot [color=black, mark=*, discard if not={model}{thinet}] table [x=gflops, y=prec1] {combined.dat};
      \addlegendentry{ThiNet\cite{luo2017thinet}};
      
      \addplot [color=brown, mark=*, discard if not={model}{geom}] table [x=gflops, y=prec1] {combined.dat};
      \addlegendentry{Geom\cite{he2019filter}};

      \addplot [color=cyan, mark=*] coordinates {(4.028, 76.13)};
      \addlegendentry{ResNet-50\cite{he2016resnet}};

      \draw [dashed, color=cyan] (axis cs:\pgfkeysvalueof{/pgfplots/xmin}, 76.13) -- (axis cs:\pgfkeysvalueof{/pgfplots/xmax}, 76.13);
\end{axis}
\end{tikzpicture}
\caption{Pruning results for ResNet-50\newline}
\label{fig:resind}
\end{subfigure}
\caption{Selected experimental results for ImageNet.}\label{fig:data}
\end{figure}
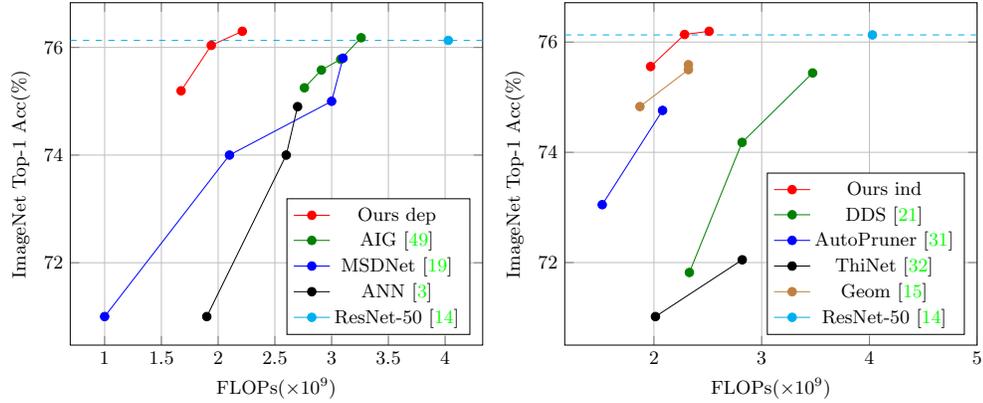

Conditional computation results are shown in Figure~\ref{fig:resdep} and Table \ref{tab:conditional}. We find that we can skip 45-52\% of the FLOPs from the baseline with comparable or even slightly better accuracy. ResNet-50 achieves a Top-1 accuracy of 76.13 with 4.028 FLOPs ($\times 10^9$). With target rate $t=.5$ we have a small improvement in accuracy (Top-1 accuracy of 76.3) at 2.21 FLOPs, which is 45\% fewer. At $t=.4$ we have a small loss in accuracy (76.04) at  1.94 FLOPs, which is 52\% fewer. The figures also show comparisons with AIG \cite{aig} (which is at the layer, rather than filter, granularity); we achieve a slightly higher accuracy with over 30\% fewer FLOPs.

\begin{table*}
\begin{tabular}{|c|cccccc|}
\hline
\pbox{1cm}{Depth} & Model & \pbox{1.6cm}{\centering Baseline\newline top-1\newline acc.($\%$)} &  \pbox{1.5cm}{\centering Accelerated\newline top-1\newline acc. ($\%$)} & \pbox{1.6cm}{\centering Top-1\newline acc $\downarrow$} & \pbox{1.4cm}{\centering Top-5\newline acc $\downarrow$} & \pbox{1.2cm}{FLOPs\newline$\downarrow$ ($\%$)} \\ \hline\hline
 & Geom\cite{he2019filter} (only 30\%) & 70.28 & 68.34 & 1.94 & 1.10 & 41.8 \\
18 & Geom\cite{he2019filter} (mix 30\%) & 70.28 & 68.41 & 1.87 & 1.15 & 41.8 \\
 & Ours ind  & 70.28 & {\bf 68.88} & {\bf 1.40} & {\bf 0.97} & {\bf 43.9} \\
\hline\hline
 & Geom\cite{he2019filter} (only 30\%) & 73.92 & 72.54 & 1.38 & \bf0.49 & 41.1 \\
34 & Geom\cite{he2019filter} (mix 30\%) & 73.92 & 72.63 & 1.29 & 0.54 & 41.1 \\
 & Ours ind  & 73.92 & \bf72.78 & \bf1.14 & 0.69 & \bf51.1 \\
\hline\hline
 & DDS-41& 76.13& 75.44& 0.69 & 2.25& 13.77 \\
 & DDS-32& 76.13& 74.18& 1.95& 1.04 & 30.0 \\
 & DDS-26& 76.13& 71.82& 4.31 & 0.29 & 42.17 \\
 & ThiNet-70& 72.88&72.04& 0.84& 3.08 & 36.7 \\
50 & AutoPruner $0.5$& 76.13& 74.76& 1.37 & 0.79 & 48.36 \\
 & Geom (only 30\%) & 76.15 & 75.59 & 0.56 & 0.24 & 42.2 \\
 & Geom (mix 30\%) & 76.15 & 75.50 & 0.65 & 0.21 & 42.2 \\
 & Geom (only 40\%) & 76.15 & 74.83 & 1.32 & 0.55 & \bf53.5 \\
 & Ours ind $t=.5$& 76.13&\bf 76.20&\bf-0.07 & \bf-0.2 & 37.68 \\
 & Ours ind $t=.4$& 76.13&76.14 &-0.01 & -0.04 & 43.39 \\
 & Ours ind $t=.3$& 76.13&75.56 & 0.56&  0.36 & 51.3 \\ \hline
\end{tabular}
\caption{Comparison of pruned ResNet on ImageNet-2012. Acc $\downarrow$ is the decrease in accuracy between the accelerated model and the baseline mode; lower is better. Baseline numbers are listed because different researchers' implementations vary; note that our compressed model for ResNet-34 is smaller than that of Geom\cite{he2019filter}. }\label{tab:pruning} 
\end{table*}

\begin{table}
    \centering
    \begin{tabular}{c|cccc}
\toprule
Variant&FLOP \%& \pbox{2.2cm}{\centering Baseline\newline top-1 acc (\%)} & \pbox{2.2cm}{\centering Accelerated\newline top-1 acc (\%)} & Top-1 $\Delta$ \\
\midrule\midrule
\multicolumn{5}{c}{Conditional computation on ResNet110} \\
\midrule
%ResNet110 & 2.52 & 93.39  & \\
%AIG-110 $t=.8$ & 82$\%$ & $93.39 \rightarrow 94.24$ & $1\% \uparrow$ \\
%Ours dep $t=.6$ & 65$\%$ & $93.39 \rightarrow 94.36$ & $1\% \uparrow$\\
AIG-110 $t=.8$ & 82$\%$ & $93.39$ & $94.24$ & $1\% \uparrow$ \\
Ours dep $t=.6$ & 65$\%$ & $93.39$ & $94.36$ & $1\% \uparrow$\\

%Filter Dep PB & 60\% & 93.39 \rightarrow 93.9 & $0.6\%$ \uparrow\\
\midrule
\multicolumn{5}{c}{Pruning on ResNet-56}\\
\hline
%AMC & $50\%$ & $92.8 \rightarrow 91.9$  & $1.0\% \downarrow$\\
%Ours ind $t=.5$ & $50\%$ & $93.86 \rightarrow 93.31$ & $0.4\% \downarrow$ \\
AMC & $50\%$ & $92.8$ & $91.9$  & $1.0\% \downarrow$\\
Ours ind $t=.5$ & $50\%$ & $93.86$ & $93.31$ & $0.4\% \downarrow$ \\

    \end{tabular}
    \caption{CIFAR-10 results. The FLOPs is reported as a percentage of the original model and accuracy is reported for the baseline and accelerated models. Note that our ResNet-56 baseline is more accurate than AMC's ResNet-56 baseline.}\label{tab:cifar}
\end{table}

\begin{table}
    \begin{tabular}{c|cccc}
    \toprule
    Model & \pbox{2cm}{\centering Baseline\newline FLOPs ($10^9$)} & \pbox{2cm}{\centering Accelerated\newline FLOPs ($10^9$)} & FLOPs $\Delta$ & Top-1 acc. (\%) \\ \midrule\midrule
    AIG-50 $t=0.6$ & 4.028 & 3.08 & 76.5\%& 75.78 \\
    Ours data-dep. layer $t=0.5$ & 4.028 &  2.72 & 67.5\% & 75.78 \\
    Ours data-dep. filter $t=0.4$ & 4.028 & 1.94 & 48.1\%& 76.07 \\
    \end{tabular}\\
    \caption{Layer vs Filter granularity for gating. FLOPs $\Delta$ is calculated from baseline ResNet50 architecture (lower is better).} \label{tab:ablationfilterlayer}
\end{table}

\subsection{Results on CIFAR-10}

We report results on CIFAR-10 on several architectures and compare to other techniques; see Figure~\ref{tab:cifar}. Using conditional computation, we obtain higher accuracy on ResNet-110, $94.36$, with 65\% fewer FLOPs. Compared to AIG, we obtain higher accuracy with 20.7\% fewer FLOPs.  Using pruning on ResNet-56, we can reduce the number of FLOPs by 50\% with only a small decrease in final accuracy, $93.31$. Compared with AMC, we have a smaller decrease in accuracy at the same FLOPs reduction. Additional results are included in the supplemental.

\subsection{Analysis and ablation studies}

\subsubsection{Filter vs layers}
\label{secsub:ablationfiltervslayer}

Our proposed techniques can be used on a layer basis; our per-batch activation loss, in combination with the Gumbel, still provides strong performance. In general, operating at filter granularity rather than layers provides a substantial boost: roughly 20\% improvement in FLOPs at the same accuracy. Results are shown in Table \ref{tab:ablationfilterlayer}. For pruning (data-independent gates), moving from layer to filter granularity results in a 28\% improvement in FLOPs for a similar accuracy.  For conditional computation (data-dependent gates), we can do an even more detailed ablation study since the primary difference between AIG\cite{aig} and our result are the batch activation loss and the filter granularity. Overall, batch activation loss provides approximately a 12\% boost over AIG and filter granularity provides an additional 27\% improvement over the layer-based version of our technique.

\subsubsection{Specialization results}

\begin{figure}
\includegraphics[width=0.6\textwidth]{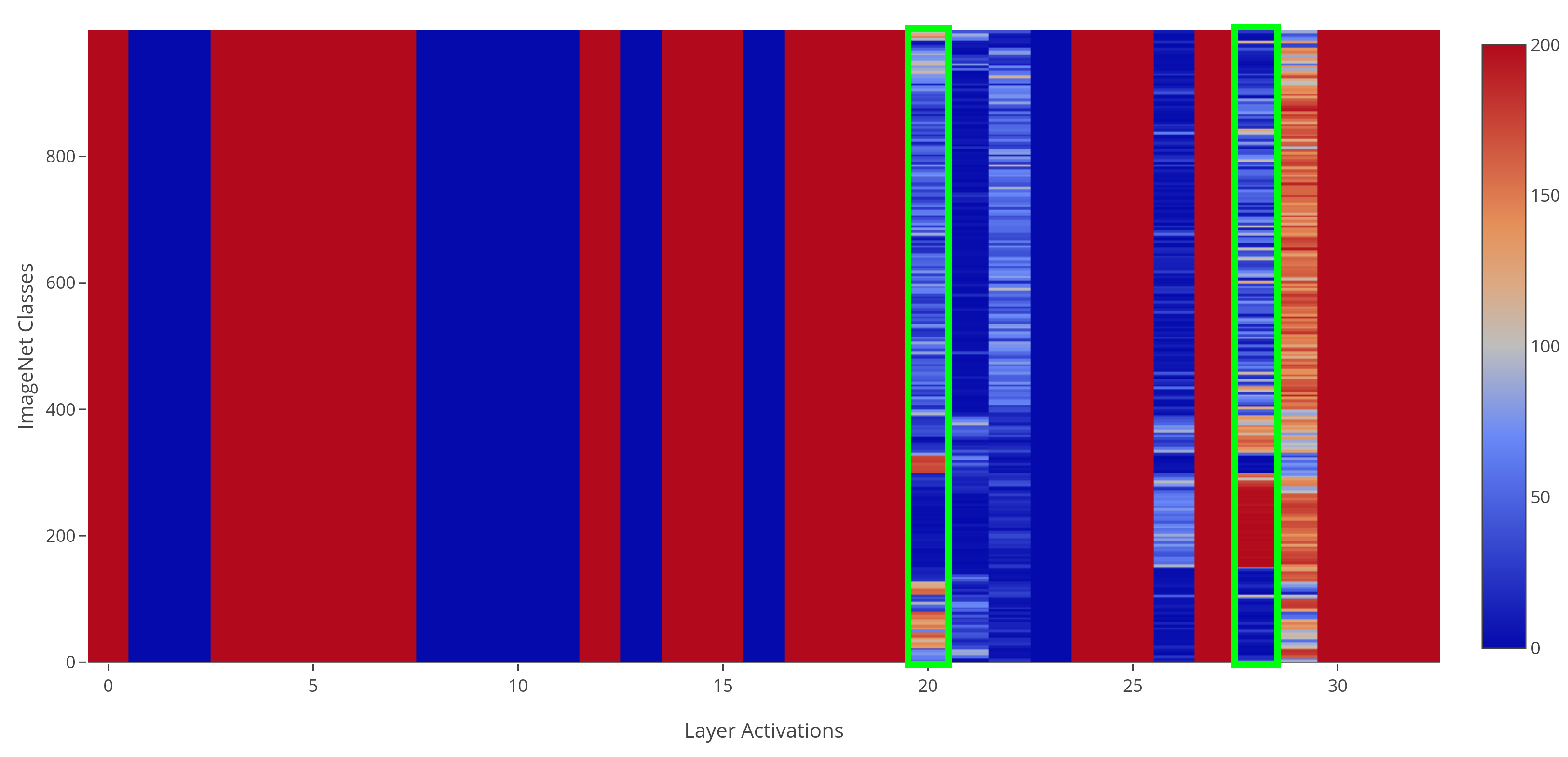} \caption{Gate activations for a layer-based conditional computation model with our batch-activation loss. Two of the layers are highlighted in green. These layers depict a higher level of specialization than those shown in AIG; the first layer runs on cats and dogs, while the second layer runs primarily on fish and lizards.}\label{fig:heatmap}
\end{figure}

In Figure \ref{fig:heatmap}, we show the gate activations for our layer-based data-dependent model. We observe higher levels of specialization than AIG. While AIG's model specializes primarily on manmade vs non-manmade objects, we specialize on more granular category types. The first layer runs only on cats and dogs, while the second layer runs primarily on fish and lizards. Note, that the specialization shown in AIG is a direct result of the chosen target rate. Their target rate of $t=0.5$ causes each layer in their model to specialize on one half of the dataset. Since our target rate is for the entire network, each layer in our model is able to specialize on whatever subset it wants. In fact, our specialized layers (cats and dogs; lizards and fish) suggest that working on smaller, more specific subsets can help the model's accuracy.

\def\mywid{3.55in}

\section{DenseNet extensions}

There are a number of natural extensions to our work that we have explored. Here, we focus on the use of probabilistic gates to provide an early exit, when the network is certain of the answer. We are motivated by MSDNet \cite{msdnet}, which investigated any-time classification. We explore early exit on both ResNet and DenseNet; however, consistent with \cite{msdnet}, we found that ResNet tends to degrade with intermediate classifiers while DenseNet does not. Following \cite{branchnet} we place gates and intermediate classifiers at the end of each dense block. At each gate, the network makes a decision as to whether the instance can be successfully classified.  Results are shown in Table~\ref{tab:early_exit_table}. These early exit gates make good decisions regarding which instances to classify early. More details are in the supplementals.

\begin{table}
    \centering
    \small
    \ignore{
    \begin{tabular}{c|ccc}
    \toprule
     & Block 1 & Block 2 & Final Block \\
     \midrule \midrule
    Images classified & 28.71\% & 11.56\% & 59.73\% \\
    Acc (all images) &  81.36 & 93.35 & 94.19 \\
    Acc (chosen images) &  96.37 & 98.53 & 92.63 \\
    \end{tabular}\\
    }
    \begin{tabular}{c|ccc}
    \toprule
     & \pbox{2cm}{\centering Images\newline selected (\%)} & \pbox{4cm}{\centering Top-1 acc. (\%) \newline (all images)} & \pbox{4cm}{\centering Top-1 acc. (\%)\newline (selected images)} \\
     \midrule \midrule
    Block 1 & 28.71\% & 81.36 & 96.37 \\
    Block 2 &  11.56\%  & 93.35 & 98.53  \\
    Final Block & 59.74\%  & 94.19 & 92.63 \\
    \end{tabular}\\
    \caption{DenseNet on CIFAR-10 with early exit. For early classifiers, the accuracy on the images that the network selects is higher than the accuracy on all images. This suggests that the gates are learning to recognize ``easy'' examples.}
\label{tab:early_exit_table}
\end{table}

\section{Conclusion}
We show an end-to-end trainable system for selecting channels using Gumbel soft-max. We propose a single framework that can handle both pruning and conditional computation at the channel level. Our novel batch loss, combined with the Gumbel trick for making categorical gate decisions, shows strong quantitative speedups over the baseline, improving the FLOPS-accuracy Pareto frontier.

$\newline$
{\bf Acknowledgments.}  We thank Andreas Veit and Serge Belongie for their invaluable insights on AIG and several reviewers for helpful comments. This work was generously supported by Google Cloud, without whose help it could not have been completed. It was funded by NSF grant IIS-1447473, by a gift from Sensetime, and by a Google Faculty Research Award.

\clearpage

\bibliographystyle{splncs04}
\bibliography{gates}

\end{document}